\documentclass[sigconf]{acmart}
\AtBeginDocument{%
  }

\setcopyright{acmlicensed}
\copyrightyear{2026}
\acmYear{2026}
\acmDOI{XXXXXXX.XXXXXXX}
\acmConference[KDD 2026]{32nd SIGKDD Conference on Knowledge Discovery and Data Mining, 2026}{Aug 09--13,
  2026}{Jeju, South Korea}
\acmISBN{978-1-4503-XXXX-X/2018/06}

\usepackage{multirow}
\usepackage{algorithm}
\usepackage{algorithmic}
\usepackage{tabularx}
\usepackage{booktabs}
\usepackage{enumitem}
\begin{document}

\title{Is Meta-Path Attention an Explanation? Evidence of Alignment and Decoupling in Heterogeneous GNNs}





\author{Maiqi Jiang}
\affiliation{%
  \institution{William \& Mary}
  \city{Williamsburg}
  \state{Virginia}
  \country{USA}
  }
  \email{mjiang04@wm.edu}

\author{Noman Ali}
\affiliation{%
  \institution{Indian Institute of Technology}
  \city{Jodhpur}
  \country{India}
  }
\email{21f1004271@ds.study.iitm.ac.in}

\author{Yiran Ding}
\affiliation{%
  \institution{Brown University}
  \city{Providence}
    \state{Rhode Island}
  \country{USA}
  }
\email{yiran_ding@alumni.brown.edu}

\author{Yanfu Zhang}
\authornote{Corresponding author.}
\affiliation{%
  \institution{William \& Mary}
  \city{Williamsburg}
  \state{Virginia}
  \country{USA}
  }
\email{yzhang105@wm.edu}

\renewcommand{\shortauthors}{Jiang et al.}


\begin{abstract}
Meta-path-based heterogeneous graph neural networks aggregate over meta-path-induced views, and their semantic-level attention over meta-path channels is widely used as a narrative for ``which semantics matter.'' We study this assumption empirically by asking: \emph{when does meta-path attention reflect meta-path importance, and when can it decouple?} A key challenge is that most post-hoc GNN explainers are designed for homogeneous graphs, and naive adaptations to heterogeneous neighborhoods can mix semantics and confound perturbations. To enable a controlled empirical analysis, we introduce \textbf{MetaXplain}, a meta-path-aware post-hoc explanation protocol that applies existing explainers in the native meta-path view domain via (i) view-factorized explanations, (ii) schema-valid channel-wise perturbations, and (iii) fusion-aware attribution, without modifying the underlying predictor. We benchmark representative gradient-, perturbation-, and Shapley-style explainers on ACM, DBLP, and IMDB with HAN and HAN-GCN, comparing against xPath and type-matched random baselines under standard faithfulness metrics. To quantify attention reliability, we propose \textbf{Meta-Path Attention--Explanation Alignment (MP-AEA)}, which measures rank correlation between learned attention weights and explanation-derived meta-path contribution scores across random runs. Our results show that meta-path-aware explanations typically outperform random controls, while MP-AEA reveals both high-alignment and statistically significant decoupling regimes depending on the dataset and backbone; moreover, retraining on explanation-induced subgraphs often preserves, and in some noisy regimes improves, predictive performance, suggesting an explanation-as-denoising effect. 
\end{abstract}





\maketitle

\section{Introduction}
Graph Neural Networks (GNNs) have demonstrated significant success in processing graph-structured data \cite{social1, social2,ecommerce1,Biological1}, and heterogeneous GNNs (HeteroGNNs) extend this capability to heterogeneous graphs by incorporating the multi-typed nature of nodes and edges. Many HeteroGNNs, such as HAN \cite{han} and MAGNN \cite{magnn}, utilize \textit{meta-paths}, sequences of node and edge types that capture meaningful relationships within the network. These meta-paths serve as powerful tools for representing task-relevant semantics within the graph structure.

\emph{Exploration goal: when is meta-path attention explanatory?}
Meta-path-based HeteroGNNs are often interpreted through the lens of \emph{semantic-level attention} over meta-path channels: for example, in HAN, the learned attention weights over meta-path views are frequently treated as a post-hoc explanation of ``which semantics matter.'' 
However, a growing literature in both NLP and graph learning shows that attention weights need not reflect feature importance~\cite{AttentionisnotExplanation, TowardsTransparent, attentionForExplainInGNN}. 
This raises a basic but unresolved question for heterogeneous graph learning:
{Does semantic-level attention reliably track meta-path importance, and if not, when does it decouple?}
This paper treats that question as an empirical one.

\emph{Why existing explainers can mislead under heterogeneity.}
Post-hoc explainability for homogeneous-graph GNNs has advanced rapidly, with gradient-, perturbation-, and search-based methods that attribute predictions to nodes, edges, or features~\cite{gradcamgnn, gnnexplainer, pgm, subgraphx, gnnshap}. 
But directly applying these explainers to meta-path-based HeteroGNNs is not straightforward: a common naive strategy collapses the heterogeneous neighborhood into a single graph and runs a homogeneous explainer. 
Such collapsing can (i) \emph{collapse semantics} by mixing edges arising from different meta-path views into indistinguishable connectivity, (ii) \emph{confound perturbations} because editing a shared structural element can simultaneously disrupt multiple semantic channels, and (iii) \emph{distort attribution under fusion} because cross-meta-path aggregation can suppress or cancel contributions at the output even when a meta-path is locally important.
As a result, both attention-based narratives and post-hoc explanations can be difficult to interpret at the meta-path (semantic-channel) level and may fail basic faithfulness checks.

\emph{A meta-path-consistent protocol for post-hoc explanation.}
To enable a controlled empirical study of meta-path importance and attention reliability, we introduce \textbf{MetaXplain}, a {meta-path-aware post-hoc explainability protocol} for meta-path-based HeteroGNNs. 
MetaXplain is not a new predictor; rather, it standardizes how existing explainers are applied in the {native computation domain} of meta-path-based models. 
The key idea is to treat the meta-path-induced local view set
$\mathcal{G}(v)=\{G^{(m)}_v\}_{m\in\mathcal{M}}$
as the explanation domain, and to enforce three {consistency conditions} that prevent invalid conclusions:
(\textbf{i}) \emph{view-factorized} explanation objects indexed by meta-path views,
(\textbf{ii}) \emph{channel-wise, schema-valid} perturbations or masks restricted within each view,
and (\textbf{iii}) \emph{fusion-aware} attribution that summarizes meta-path contributions without collapsing semantics.
This protocol ``lifts'' a broad class of homogeneous-graph explainers into meta-path-induced view spaces without modifying the underlying trained model.

\emph{Diagnostic: measuring attention--importance agreement across runs.}
To directly test whether semantic attention is a faithful proxy for meta-path importance, we propose
\textbf{Meta-Path Attention--Explanation Alignment (MP-AEA)}, which quantifies the agreement between
(i) a model's semantic attention distribution over meta-paths and
(ii) explanation-derived meta-path contribution scores.
MP-AEA measures rank correlation between these two quantities across random runs, turning the qualitative question
``is attention explanatory?'' into a reproducible diagnostic.

\emph{Empirical study and key observations.}
We instantiate MetaXplain with representative explainers spanning gradient-based, perturbation-based, and Shapley-style families
(Grad~\cite{grad,gradcamgnn}, GNNExplainer~\cite{gnnexplainer}, PGM-Explainer~\cite{pgm}, GraphSVX~\cite{graphsvx}, and GNNShap~\cite{gnnshap}), 
and benchmark on ACM, DBLP, and IMDB with HAN and HAN-GCN, comparing against xPath~\cite{xpath} and type-matched random baselines under standard faithfulness metrics.
Across datasets and backbones, lifted explainers typically outperform random baselines, but performance and stability are backbone-dependent. 
Most importantly, MP-AEA reveals that semantic attention can be reliable in some settings yet statistically decoupled from meta-path importance in others.
Finally, we find that retraining on explanation-induced subgraphs often preserves, and in noisy spectral regimes can improve predictive performance, suggesting an \emph{explanation-as-denoising} effect.

Our contributions are primarily diagnostic and empirical:
\begin{itemize}[leftmargin=*, label=\textbullet]
    \item We introduce \textbf{MetaXplain}, a meta-path-consistent post-hoc explanation protocol that standardizes how diverse homogeneous-graph explainers can be applied in the meta-path view domain (view-factorized explanations, schema-valid channel-wise perturbation, fusion-aware attribution).
    \item We propose \textbf{MP-AEA}, a diagnostic that measures whether semantic-level meta-path attention tracks explanation-derived meta-path contributions across random runs.
    \item Through experiments on ACM, DBLP, and IMDB with HAN/HAN-GCN, we report empirical findings that (i) meta-path-aware explainers improve over type-matched random baselines but can be backbone-sensitive, (ii) semantic attention can align with or decouple from meta-path importance depending on the dataset / backbone, and (iii) explanation-induced subgraphs can act as denoising and support effective retraining.
\end{itemize}

\section{Related Work}

\emph{Post-hoc explainability for homogeneous-graph GNNs.}
Post-hoc explainability for homogeneous GNNs spans (i) gradient/saliency attribution and CAM-style variants~\cite{sa,grad,gradcam,gradcamgnn}, 
(ii) mask-optimization explainers such as GNNExplainer and PGExplainer~\cite{gnnexplainer,pge}, 
(iii) probabilistic/causal approaches (e.g., PGM-Explainer)~\cite{pgm}, and 
(iv) Shapley-style attribution (e.g., GraphSVX, GNNShap)~\cite{graphsvx,gnnshap}. 
Most of these methods assume a single unlabeled edge space and a homogeneous neighborhood, so their explanation variables (masks, perturbations, players) are not directly aligned with the \emph{semantic channel decomposition} induced by meta-path views in heterogeneous graphs. 
In this paper, our goal is not to propose a new homogeneous explainer, but to enable a controlled \emph{empirical study} of meta-path importance by applying existing explainers in a meta-path-consistent way.

\emph{Explainability for heterogeneous graphs.}
Explainability for heterogeneous graphs is less mature and often model- or task-specific. 
PaGE-Link~\cite{pagelink} explains link prediction via influential paths, while other work derives symbolic or causal criteria under heterogeneity constraints~\cite{ce,CFHGExplainer,hencex}. 
Most closely related, xPath~\cite{xpath} generates type-aware explainable paths for heterogeneous predictions. 
These approaches highlight the importance of respecting schema semantics, but they do not directly provide a standardized protocol for analyzing meta-path-based pipelines that explicitly operate on meta-path-induced views (e.g., HAN~\cite{han}).
Our work is complementary: we focus on meta-path-based HeteroGNNs and provide a meta-path-aware post-hoc protocol that standardizes how diverse existing explainers can be applied to the meta-path view domain, enabling apples-to-apples comparisons and downstream diagnostics of semantic-channel importance.

\emph{Is attention explanatory? Diagnostics and reliability.}
Attention is frequently used as an explanation mechanism, yet in NLP it is neither necessary nor sufficient in general~\cite{AttentionisnotExplanation,TowardsTransparent}, with follow-up work studying conditions under which attention aligns with feature importance~\cite{AttentionisnotnotExplanation}. 
Similar concerns have been raised for attention mechanisms in graph models~\cite{attentionForExplainInGNN}. 
Despite the prevalence of \emph{semantic-level meta-path attention} in models such as HAN, there is limited systematic evidence on when such attention reliably reflects meta-path importance rather than acting as a convenient but potentially misleading narrative. 
Motivated by this gap, our paper contributes a simple diagnostic, \textbf{MP-AEA}, that measures attention--importance agreement across random runs using explanation-derived meta-path contribution scores. 
This diagnostic supports our central exploration question: when does meta-path attention align with, or decouple from, meta-path importance?

\begin{figure*}[ht]
    \centering
    \includegraphics[scale=0.6]{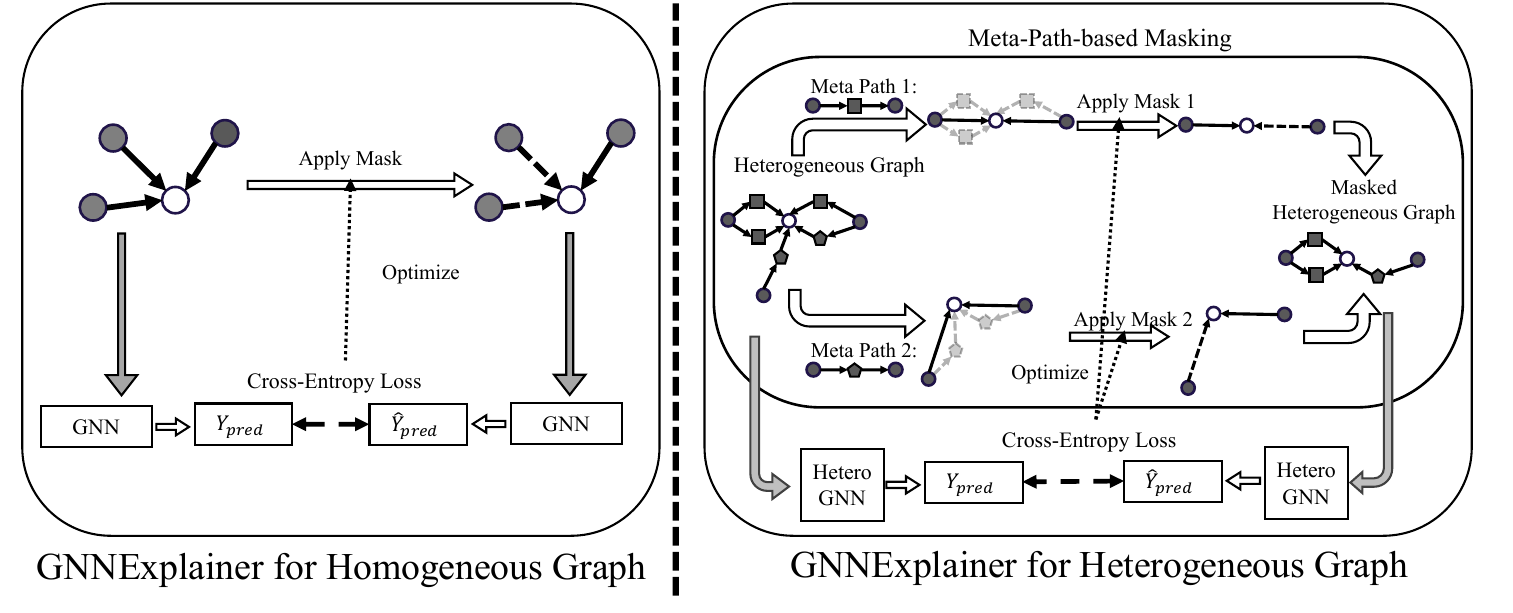}
    \caption{\textbf{MetaXplain lifts GNNExplainer to the meta-path view domain.}
}
    \label{main_fig}
\vspace{-1em}
\end{figure*}

\section{Preliminaries}
\label{sec:prelim}

A \textbf{heterogeneous graph} models multiple node and/or edge types to capture complex real-world semantics. Formally, let a typed graph be denoted by $\mathcal{G}=(\mathcal{V},\mathcal{E},\mathcal{T},\mathcal{R})$, where $\mathcal{T}$ is the set of node types and $\mathcal{R}$ is the set of relation (edge) types, with type mappings $\phi:\mathcal{V}\!\to\!\mathcal{T}$ and $\psi:\mathcal{E}\!\to\!\mathcal{R}$. The graph is heterogeneous if $|\mathcal{T}|>1$ or $|\mathcal{R}|>1$; otherwise, it is homogeneous.

A meta-path is a typed relation sequence defined on the schema, capturing a semantic pattern of connectivity.
Formally, a meta-path $m$ can be written as $m:\quad t_1 \xrightarrow{r_1} t_2 \xrightarrow{r_2} \cdots \xrightarrow{r_\ell} t_{\ell+1},$
where $t_i\in\mathcal{T}$ and $r_i\in\mathcal{R}$.
Two nodes $u$ and $v$ are meta-path-connected by $m$ if there exists an instance path in $\mathcal{G}$ whose node/edge types match the sequence $m$.

Given a meta-path $m\in\mathcal{M}$, we denote the \emph{meta-path-induced view} (or meta-path graph) as $G^{(m)}=(\mathcal{V},\mathcal{E}^{(m)})$, where $(u,v)\in\mathcal{E}^{(m)}$ if $u$ and $v$ are connected by at least one instance path following $m$.
This construction is widely used in meta-path-based heterogeneous learning to provide a semantic-specific neighborhood for aggregation.

For a target node $v$, we define its \emph{local meta-path computation graphs} as
\begin{equation}
\mathcal{G}(v)=\{G^{(m)}_v \mid m\in\mathcal{M}\},
\end{equation}
where $G^{(m)}_v$ is typically an ego-subgraph (e.g., $k$-hop neighborhood) of $v$ extracted from $G^{(m)}$.
We treat $\mathcal{G}(v)$ as the native computation domain of meta-path-based models and, correspondingly, the native domain for post-hoc explanation in our study.

\subsection{Meta-path-based HeteroGNN Interface}
Many meta-path-based HeteroGNNs follow a two-stage computation: (i) \emph{within-meta-path} aggregation to compute channel-wise embeddings and (ii) \emph{cross-meta-path} fusion to produce the final representation and prediction.
We abstract this family via the following interface:
\begin{align}
\mathbf{h}^{(m)}_v &= g^{(m)}\!\left(G^{(m)}_v, X\right), \qquad \forall m\in\mathcal{M}, \label{eq:mp-encode}\\
f(v) &= \mathrm{Dec}(\mathbf{h}_v), \mathbf{h}_v = \mathrm{Fuse}\!\left(\{\mathbf{h}^{(m)}_v\}_{m\in\mathcal{M}}\right), \label{eq:mp-predict}
\end{align}
where $X$ denotes node features (optionally type-specific), $\mathrm{Fuse}(\cdot)$ is a semantic fusion operator, and $\mathrm{Dec}(\cdot)$ is a task head.
For instance, HAN~\cite{han} realizes $\mathrm{Fuse}(\cdot)$ with semantic-level attention over meta-path channels, while SeHGNN~\cite{sehgnn} uses a simplified neighbor aggregation with a semantic fusion module (e.g., transformer-style fusion) over meta-path-derived representations.

\subsection{Post-hoc Explainability Setup}
We focus on \emph{post-hoc} explanation: given a trained predictor $f$ and a specific prediction for target $v$, the goal is to output an explanation object $\mathcal{E}_v$ that is (i) compact and (ii) faithful to $f$.
In homogeneous graphs, model-agnostic explainers such as GNNExplainer seek a small subgraph and a subset of node features that maximize fidelity to the prediction.
In our setting, the explanation space is meta-path-factorized: explanations may be defined per view $G^{(m)}_v$ (e.g., edge masks or selected structures).
When needed for semantic interpretation, we summarize each $E^{(m)}_v$ into a nonnegative scalar \emph{meta-path contribution score}, which we later compare against the model’s semantic attention in our diagnostic (MP-AEA).

\section{A Meta-Path-Consistent Explanation Protocol}

\noindent\textbf{Goal.}
Our objective is to enable a controlled empirical study of meta-path importance and the reliability of semantic-level meta-path attention.
MetaXplain standardizes how existing post-hoc explainers are applied in the meta-path view domain to avoid invalid conclusions caused by heterogeneity.

\subsection{MetaXplain: Consistency conditions for meta-path-aware post-hoc explanation}
\label{sec:framework}

\noindent\textbf{Context.}
As defined in Section~\ref{sec:prelim}, a meta-path-based HeteroGNN takes as input the local meta-path views
$\mathcal{G}(v)=\{G^{(m)}_v\}_{m\in\mathcal{M}}$ and computes meta-path-specific embeddings $\{\mathbf{h}_v^{(m)}\}$
followed by a fusion $\mathbf{h}_v=\mathrm{Fuse}(\{\mathbf{h}_v^{(m)}\})$ and prediction $f(v)=\mathrm{Dec}(\mathbf{h}_v)$
(Equations~\eqref{eq:mp-encode}--\eqref{eq:mp-predict}).
Naively applying homogeneous explainers on a collapsed graph can yield confounded explanations due to:
(i) \emph{semantic collapse}, (ii) \emph{relation confusion} under perturbation, and (iii) \emph{fusion-induced suppression} in attribution. We therefore enforce three consistency conditions below.

\subsubsection{C1: View-Factorized Explanation Domain}
\label{method:principle1}

\paragraph{Condition.}
Under MetaXplain, explanations are defined on the meta-path view set
$\mathcal{G}(v)=\{G^{(m)}_v\}_{m\in\mathcal{M}}$ rather than a single collapsed graph.
Equivalently, its output decomposes as
\begin{equation}
\label{eq:p1_def}
\mathcal{E}_v \;=\; \{E^{(m)}_v\}_{m\in\mathcal{M}},
\end{equation}
where each $E^{(m)}_v$ is supported on the corresponding view $G^{(m)}_v$.
This condition prevents \emph{semantic collapse} and ensures explanations remain interpretable at the semantic-channel level.

\emph{Applying a homogeneous explainer under C1.} 
Let a standard explainer be $\Phi$ that outputs an explanation for a (homogeneous) local graph input:
$E_v = \Phi(f, G_v, X)$.
We define its meta-path lifting as
\begin{equation}
\label{eq:p1_lift}
\Phi_{\textsc{mp}}(f, \mathcal{G}(v), X) \;:=\; \{E^{(m)}_v\}_{m\in\mathcal{M}},
\end{equation}
i.e., the explainer now optimizes / searches over a factorized explanation space aligned with the meta-path views.

\subsubsection{C2: Channel-Wise and Schema-Valid Perturbation}
\label{method:principle2}

\paragraph{Condition.}
A perturbation mechanism is \emph{meta-path-consistent} if it is (i) \emph{channel-wise} (independent across $m$)
and (ii) \emph{schema-valid} (restricted to view $G_v^{(m)}$ and type consistent).
Formally, perturbations take the form $\{\delta^{(m)}\}_{m\in\mathcal{M}}$ where each $\delta^{(m)}$ acts only within $G^{(m)}_v$.

\emph{Math (disentangled masks and optional semantic gate).}
For mask-based explainers, we replace a single mask by per-view masks
\begin{equation}
\label{eq:p2_masks}
M=\{M^{(m)}\}_{m\in\mathcal{M}}, \qquad M^{(m)} \in [0,1]^{|\mathcal{E}(G^{(m)}_v)|},
\end{equation}
and define a masked view $\widetilde{G}^{(m)}_v = G^{(m)}_v \odot M^{(m)}$ (masking only edges in $G^{(m)}_v$). This condition prevents \emph{confounded perturbations} in which removing a shared structural element simultaneously disrupts multiple meta-path semantics.

Optionally, we introduce a meta-path gate $m_{\mathrm{sem}}\in[0,1]^{|\mathcal{M}|}$ and couple it to instance-level masks:
\begin{equation}
\label{eq:p2_gate}
\widetilde{M}^{(m)} := m_{\mathrm{sem}}^{(m)}\cdot M^{(m)}, \qquad \widetilde{G}^{(m)}_v = G^{(m)}_v \odot \widetilde{M}^{(m)}.
\end{equation}
The explainer then optimizes its original objective (e.g., fidelity with sparsity regularization) over $\{\widetilde{M}^{(m)}\}$.

\subsubsection{C3: Fusion-Aware Attribution (Model-Agnostic)}
\label{method:principle3}

\paragraph{Condition.}
An attribution is \emph{fusion-aware} if it measures the contribution of each semantic channel \emph{before}
cross-channel fusion, and aggregates channel-wise contributions to explain $f(v)$.

\emph{Two practical options.}
Using the interface in Equations~\eqref{eq:mp-encode}--\eqref{eq:mp-predict}, define a channel contribution score
$s_v^{(m)}$ and an aggregation $\mathrm{Agg}(\cdot)$:
\begin{equation}
\label{eq:p3_agg}
\mathrm{Score}(v) = \mathrm{Agg}\big(\{s_v^{(m)}\}_{m\in\mathcal{M}}\big).
\end{equation}
This avoids \emph{fusion-induced attribution artifacts}, where contributions can appear suppressed after $\mathrm{Fuse}(\cdot)$
despite being locally important within a view.

\textit{(A) Gradient-based (late-fusion gradients).}
Fusion-aware gradients can be computed either on the pre-fusion channel embeddings $\mathbf{h}_v^{(m)}$ or on the channel-specific inputs (e.g., features in view $m$), depending on the model accessibility:
\begin{equation}
\label{eq:p3_grad}
s_{v,\textsc{grad}}^{(m)} \in 
\left\{
\left\|\nabla_{\mathbf{h}_v^{(m)}} \mathcal{L}\right\|,
\;
\left\|\nabla_{X^{(m)}} \mathcal{L}\right\|
\right\}.
\end{equation}

\textit{(B) Gradient-free (fusion-aware occlusion / conditional fidelity).}
To remain post-hoc and model-agnostic, we estimate channel importance by occluding only channel $m$ while holding all other channels fixed:
\begin{equation}
\label{eq:p3_occ}
s_{v,\textsc{occ}}^{(m)} \;=\;
D\!\left(
f\!\left(v;\{G^{(k)}_v\}_{k\in\mathcal{M}}\right),
f\!\left(v;\{G^{(k)}_v\}_{k\neq m},\ \varnothing^{(m)}\right)
\right),
\end{equation}
where $\varnothing^{(m)}$ is a schema-valid null view (e.g., $\widetilde{M}^{(m)}=\mathbf{0}$) and $D(\cdot,\cdot)$ is a logit/probability divergence.
Moreover, within-channel structure attribution follows by substituting $\varnothing^{(m)}$ with the masked view $\widetilde{G}^{(m)}_v$
from C2 and measuring the conditional fidelity drop.

\subsection{Explainers used in our empirical study}
\label{sec:instantiations}

We evaluate representative post-hoc explainers spanning gradient-, perturbation-, and Shapley-style families.
Under MetaXplain, each explainer is applied in the meta-path view domain and constrained by C1--C3.
We emphasize that these are representative choices for our empirical study; the protocol is compatible with other post-hoc explainers.

\paragraph{Local view extraction and notation.}
Given a target node $v$, we extract a $k$-hop ego neighborhood {within each meta-path view} $G^{(m)}$
and re-index to a compact local graph $G^{(m)}_v$ for efficient explanation.
This yields a list of sparse local adjacencies $\{A^{(m)}\}_{m=1}^{M}$ (one per meta-path) and a
local feature matrix $X$ shared across views.
When an explainer requires feature perturbations conditioned on a specific view, we use per-view
feature copies $\{X^{(m)}\}$ while keeping the adjacency list fixed.

\textbf{Gradient heatmap (Grad).}
Under C1, Grad produces a view-indexed node-importance mask per meta-path view.
For a target node $v$ and its predicted class $c^\star$, we compute gradients of the node-level loss
with respect to node features and take the $\ell_2$ norm across feature dimensions as a node score.
Concretely, for node $i$ in view $m$,
\begin{equation}
I_{\mathrm{Grad}}^{(m)}(i)
=
\left\lVert \frac{\partial \mathcal{L}(v,c^\star)}{\partial x_i^{(m)}} \right\rVert_2.
\end{equation}
To realize {meta-path-conditioned} gradients, when the predictor supports multi-feature inputs,
we pass a {list} of feature tensors (one per meta-path) with independent gradient tracking,
while keeping the meta-path adjacency list $\{A^{(m)}\}$ fixed.
The resulting explanation is the set of node masks $\{I_{\mathrm{Grad}}^{(m)}\}_{m=1}^{M}$, one per view.

\textbf{GNNExplainer.}
Under C1--C2, we apply GNNExplainer with channel-wise edge masks restricted to each view and return the per-view mask set $\{M_A^{(m)}\}$.
Given view $m$ with local adjacency $A^{(m)}$, we optimize a continuous edge mask
$M_A^{(m)}\in\mathbb{R}^{|\mathcal{E}^{(m)}|}$ (activated by sigmoid or ReLU), and (optionally) a
feature mask $M_X\in\mathbb{R}^{d}$ shared across views.
The masked input is formed by reweighting edges within each view and masking features:
\begin{equation}
\widetilde{A}^{(m)} = A^{(m)} \odot g(M_A^{(m)}), \quad
\widetilde{X} = X \odot h(M_X),
\end{equation}
where $g(\cdot)$ is the configured edge-mask activation and $h(\cdot)$ is either identity or sigmoid.
We then minimize a prediction-preservation loss with sparsity and entropy regularization (and an optional Laplacian term):
\begin{equation}
\begin{aligned}
\min_{\{M_A^{(m)}\},\,M_X}\;&
\mathcal{L}_{\mathrm{pred}}\!\left(f(\{\widetilde{A}^{(m)}\},\widetilde{X});\,v\right)
+\lambda_{\mathrm{size}}\!\sum_{m}\|g(M_A^{(m)})\|_1 \\
&+\lambda_{\mathrm{ent}}\!\sum_{m}\mathcal{H}(g(M_A^{(m)}))
+\lambda_{\mathrm{lap}}\mathcal{L}_{\mathrm{lap}}.
\end{aligned}
\label{L_GNNExplainer}
\end{equation}
When feature masking is enabled, we add analogous size/entropy terms for $h(M_X)$.
All meta-path masks are optimized {jointly} but remain view-specific in output, yielding
$\{g(M_A^{(m)})\}_{m=1}^{M}$ as meta-path-aware edge explanations.

In the main paper we focus on Grad and GNNExplainer as representative explainers; additional perturbation- and Shapley-style instantiations (and full results) are reported in the Appendix.

\subsection{Metric: Meta-Path Attention--Explanation Alignment (MP-AEA)}
\label{sec:mp_aea}

Motivated by evidence that attention weights need not faithfully reflect feature importance~\cite{AttentionisnotExplanation,TowardsTransparent},
we propose \textbf{MP-AEA} as a diagnostic that tests whether semantic meta-path attention \emph{aligns with or decouples from} explanation-derived meta-path contributions across random runs.

\subsubsection{Meta-path contribution scores from view-factorized explanations}
Given a target node $v$, a {MetaXplain} instantiation outputs a view-factorized explanation object
$\mathcal{E}_v=\{E_v^{(m)}\}_{m\in\mathcal{M}}$ (C1), where each $E_v^{(m)}$ is supported on view $G_v^{(m)}$.
To compare against semantic attention, we summarize each view-level explanation into a nonnegative scalar
{meta-path contribution score} $s_v^{(m)}\in\mathbb{R}_{\ge 0}$: $s_v^{(m)} \;=\; \mathrm{Agg}\!\left(E_v^{(m)}\right)$, 
where $\mathrm{Agg}(\cdot)$ is an aggregation operator consistent with the explainer output type.
For node-attribution explainers (e.g., Grad, PGM-Explainer), we use $\mathrm{Agg}(E_v^{(m)})=\sum_{u\in \mathcal{V}(G_v^{(m)})} I_v^{(m)}(u)$;
for edge-attribution explainers (e.g., GNNExplainer, GNNShap), we use $\mathrm{Agg}(E_v^{(m)})=\sum_{e\in \mathcal{E}(G_v^{(m)})} w_v^{(m)}(e)$.

Optionally, to explicitly satisfy fusion-aware attribution (C3(B)), $s_v^{(m)}$ can be defined via channel occlusion:
\begin{equation}
\label{eq:mpaea_occ}
s_v^{(m)} \;=\;
D\!\left(
f\!\left(v;\mathcal{G}(v)\right),
f\!\left(v;\mathcal{G}(v)\setminus G_v^{(m)},\ \varnothing^{(m)}\right)
\right),
\end{equation}
where $\varnothing^{(m)}$ is a schema-valid null view (e.g., all-zero mask on $G_v^{(m)}$) and $D(\cdot,\cdot)$ is a logit/probability divergence.
Unless stated otherwise, we use Grad-based $E_v^{(m)}$ in our experiments.

\subsubsection{Global explanation-derived meta-path vector}
Let $\mathcal{T}$ denote the test node set.
We first normalize per-node contributions across meta-paths $\tilde{s}_v^{(m)} \;=\; \frac{s_v^{(m)}}{\sum_{m'\in\mathcal{M}} s_v^{(m')}}$.
We then aggregate to a dataset-level vector $\mathbf{S}\in\mathbb{R}^{|\mathcal{M}|}$:
\begin{equation}
\label{eq:mpaea_global}
\mathbf{S}_m \;=\; \frac{1}{|\mathcal{T}|}\sum_{v\in\mathcal{T}} \tilde{s}_v^{(m)}.
\end{equation}
$\mathbf{S}_m$ estimates the overall contribution of meta-path $m$ according to the chosen post-hoc explainer.

\subsubsection{Alignment with semantic attention across runs}
Let $\mathbf{A}\in\mathbb{R}^{|\mathcal{M}|}$ be the model's semantic attention distribution over meta-path channels
(after softmax normalization; e.g., HAN's semantic-level attention).
We train the model with $R$ random seeds, obtaining vector pairs $\{(\mathbf{A}^{(r)},\mathbf{S}^{(r)})\}_{r=1}^{R}$.
For each meta-path $m$, MP-AEA computes rank correlation between attention and explanation across runs:
\begin{align}
\text{MP-AEA}_{\tau}(m) &= \tau\big([\mathbf{A}^{(1)}_m,\dots,\mathbf{A}^{(R)}_m],\ [\mathbf{S}^{(1)}_m,\dots,\mathbf{S}^{(R)}_m]\big),\\
\text{MP-AEA}_{\rho}(m) &= \rho\big([\mathbf{A}^{(1)}_m,\dots,\mathbf{A}^{(R)}_m],\ [\mathbf{S}^{(1)}_m,\dots,\mathbf{S}^{(R)}_m]\big).
\end{align}

\paragraph{Interpretation.}
High MP-AEA indicates that semantic attention consistently tracks explanation-derived meta-path contributions across runs,
while low or insignificant MP-AEA reveals an interpretability gap where attention is decoupled from post-hoc importance.
MP-AEA measures consistency, not necessarily correctness.

\paragraph{Binary meta-paths.}
When $|\mathcal{M}|=2$, both $\mathbf{A}$ and $\mathbf{S}$ are complementary (summing to 1), so correlation magnitudes are identical for both meta-paths.
We therefore report a single MP-AEA score per model--dataset pair.

\begin{table*}[ht]
\footnotesize
\centering
\caption{Results of explanation models on HAN/HAN-GCN predictor for three datasets. Higher is better for all metrics.}
\begin{tabular}{cclccccc}
\hline
\textbf{Predictor}  & \textbf{HeteroGNN}  & \textbf{Model} & \textbf{Mask Type} & \textbf{$1 - {Fidelity}_{-}$} &
\textbf{${Fidelity}_{+}$}  & \textbf{$Macro-F1$} & \textbf{$Micro-F1$} \\
\hline
\multirow{18}{*}{\textbf{ACM}} & \multirow{9}{*}{\textbf{HAN}} 
& \textbf{GNNExplainer}& edge+feature & \textbf{88.20 $\pm$ 2.32} & $2.60 \pm 0.80$ & \textbf{84.02 $\pm$ 2.17} & \textbf{83.60 $\pm$ 2.33}\\
& & \textbf{RandomEdgeAndFeatureMask} & edge+feature & $85.80 \pm 2.56$ & \textbf{3.00 $\pm$ 0.63} & $80.63 \pm 2.21$ & $80.00 \pm 2.19$ \\
\cline{3-8}
& & \textbf{xPath}& edge 
& $81.60 \pm 3.44$ & $2.80 \pm 0.75$ & $79.77 \pm 2.94$ & $78.80 \pm 3.31$ \\
& & \textbf{GraphSVX}& edge 
& $94.60 \pm 2.15$ & $9.20 \pm 1.47$ & \textbf{87.95 $\pm$ 1.15} & \textbf{87.60 $\pm$ 1.20} \\
& & \textbf{GNNShap}& edge 
& \textbf{94.80 $\pm$ 1.33} & \textbf{10.40 $\pm$ 1.36} & $87.34 \pm 1.12$ & $87.00 \pm 1.10$\\
& & \textbf{RandomEdgeMask}& edge 
& $87.60 \pm 2.65$ & $2.20 \pm 1.47$ & $82.47 \pm 3.33$ & $81.80 \pm 3.43$ \\
\cline{3-8}
& & \textbf{Grad}& node 
& \textbf{97.60 $\pm$ 0.80} & \textbf{32.20 $\pm$ 3.92} & \textbf{91.49 $\pm$ 0.84} & \textbf{91.40 $\pm$ 0.80} \\
& & \textbf{PGM-Explainer}& node 
& $95.40 \pm 1.85$ & $15.60 \pm 5.95$ & $90.08 \pm 1.15$ & $90.00 \pm 1.10$  \\
& & \textbf{RandomNodeMask}& node 
& $89.20 \pm 2.40$ & $3.80 \pm 1.72$ & $85.87 \pm 2.02$ & $85.40 \pm 2.15$\\
\cline{2-8}
& \multirow{9}{*}{\textbf{HAN-GCN}}  & \textbf{GNNExplainer} & edge+feature 
& $60.60 \pm 8.16$ & $16.80 \pm 3.82$ & $59.75 \pm 7.08$ & $60.40 \pm 7.00$ \\
& & \textbf{RandomEdgeAndFeatureMask}& edge+feature 
& \textbf{72.40 $\pm$ 7.20} & \textbf{21.80 $\pm$ 5.42} & \textbf{69.98 $\pm$ 8.64} & \textbf{71.80 $\pm$ 6.62} \\
\cline{3-8}
& & \textbf{xPath}& edge 
& $75.60 \pm 3.72$ & $19.80 \pm 8.47$ & \textbf{77.62 $\pm$ 5.97} & $77.20 \pm 6.01$ \\
& & \textbf{GraphSVX}& edge 
& $77.40 \pm 7.00$ & $21.80 \pm 4.31$ & $76.40 \pm 7.20$ & $76.80 \pm 6.43$  \\
& & \textbf{GNNShap}& edge 
& \textbf{80.80 $\pm$ 11.37} & \textbf{24.40 $\pm$ 2.24} & $77.39 \pm 10.60$ & \textbf{78.40 $\pm$ 8.66} \\
& & \textbf{RandomEdgeMask}& edge 
& $73.80 \pm 7.11$ & $20.40 \pm 6.44$ & $72.16 \pm 7.24$ & $73.20 \pm 5.98$ \\
\cline{3-8}
& & \textbf{Grad}& node 
& \textbf{93.00 $\pm$ 5.02} & \textbf{32.40 $\pm$ 3.56} & \textbf{88.21 $\pm$ 5.10} & \textbf{88.00 $\pm$ 5.10}  \\
& & \textbf{PGM-Explainer}& node 
& $84.40 \pm 4.08$ & $28.60 \pm 5.24$ & $82.85 \pm 3.43$ & $82.60 \pm 3.26$  \\
& & \textbf{RandomNodeMask}& node 
& $75.20 \pm 7.00$ & $21.40 \pm 6.89$ & $73.55 \pm 7.78$ & $74.60 \pm 6.59$ \\
\hline
\multirow{18}{*}{\textbf{DBLP}} & \multirow{9}{*}{\textbf{HAN}} 
& \textbf{GNNExplainer} & edge+feature 
& \textbf{95.60 $\pm$ 2.06} & \textbf{1.60 $\pm$ 1.50} & \textbf{92.39 $\pm$ 2.06} & \textbf{94.80 $\pm$ 1.47}\\
& & \textbf{RandomEdgeAndFeatureMask} & edge+feature 
& $78.40 \pm 9.35$ & \textbf{1.60 $\pm$ 0.80} & $66.86 \pm 8.44$ & $77.60 \pm 10.29$\\
\cline{3-8}
& & \textbf{xPath} & edge 
& $93.00 \pm 1.90$ & $2.80 \pm 1.17$ & \textbf{94.52 $\pm$ 1.41} & $95.00 \pm 1.10$ \\
& & \textbf{GraphSVX} & edge 
& $97.00 \pm 0.63$ & $1.80 \pm 1.47$ & $92.75 \pm 0.71$ & $95.00 \pm 0.63$  \\
& & \textbf{GNNShap} & edge 
& \textbf{99.40 $\pm$ 0.49} & \textbf{7.40 $\pm$ 1.85} & $91.70 \pm 0.65$ & $94.80 \pm 0.40$ \\
& & \textbf{RandomEdgeMask} & edge 
& $96.00 \pm 1.10$ & $1.00 \pm 1.10$ & $92.78 \pm 1.66$ & \textbf{95.20 $\pm$ 1.17} \\
\cline{3-8}
& & \textbf{Grad}& node 
& $94.40 \pm 3.26$ & \textbf{8.80 $\pm$ 3.49} & $90.56 \pm 2.11$ & $92.40 \pm 2.42$\\
& &  \textbf{PGM-Explainer} & node 
& \textbf{95.00 $\pm$ 0.63} & $4.20 \pm 0.75$ & \textbf{91.78 $\pm$ 0.70} & \textbf{94.20 $\pm$ 0.40} \\
& & \textbf{RandomNodeMask} & node 
& $82.00 \pm 8.56$ & $2.00 \pm 0.63$ & $74.11 \pm 7.58$ & $81.60 \pm 8.80$\\
\cline{2-8}
& \multirow{9}{*}{\textbf{HAN-GCN}} 
& \textbf{GNNExplainer} & edge+feature 
& $71.60 \pm 1.02$ & $5.80 \pm 0.75$ & $62.00 \pm 1.82$ & $68.60 \pm 1.50$ \\
& & \textbf{RandomEdgeAndFeatureMask} & edge+feature 
& \textbf{93.20 $\pm$ 1.33} & \textbf{6.40 $\pm$ 0.49} & \textbf{87.63 $\pm$ 2.20} & \textbf{91.40 $\pm$ 1.62} \\
\cline{3-8}
& & \textbf{xPath} & edge 
& $81.20 \pm 2.79$ & $5.40 \pm 1.20$ & $73.61 \pm 2.61$ & $78.60 \pm 2.15$ \\
& & \textbf{GraphSVX} & edge 
& $90.60 \pm 1.85$ & $6.80 \pm 0.75$ & $84.37 \pm 1.52$ & $87.20 \pm 1.17$  \\
& & \textbf{GNNShap} & edge 
& \textbf{94.40 $\pm$ 0.80} & \textbf{11.60 $\pm$ 1.50} & \textbf{88.11 $\pm$ 2.95} & \textbf{91.40 $\pm$ 2.24} \\
& & \textbf{RandomEdgeMask} & edge 
& $93.80 \pm 0.40$ & $6.60 \pm 1.02$ & $87.76 \pm 2.46$ & $91.20 \pm 1.72$\\
\cline{3-8}
& & \textbf{Grad}& node 
& $93.60 \pm 0.49$ & \textbf{7.20 $\pm$ 0.75} & $84.88 \pm 0.14$ & $89.00 \pm 0.00$ \\
& & \textbf{PGM-Explainer}& node 
& \textbf{94.00 $\pm$ 1.41} & $6.80 \pm 0.75$ & \textbf{88.99 $\pm$ 2.77} & \textbf{92.20 $\pm$ 1.94} \\
& & \textbf{RandomNodeMask}& node 
& $93.60 \pm 0.49$ & $6.40 \pm 0.80$ & $87.74 \pm 1.36$ & $91.40 \pm 1.02$ \\
\hline
\multirow{18}{*}{\textbf{IMDB}} & \multirow{9}{*}{\textbf{HAN}} 
& \textbf{GNNExplainer} & edge+feature 
& \textbf{68.80 $\pm$ 2.14} & $7.20 \pm 1.72$ & \textbf{48.22 $\pm$ 4.04} & \textbf{51.20 $\pm$ 2.79} \\
& & \textbf{RandomEdgeAndFeatureMask} & edge+feature 
& $56.80 \pm 4.87$ & \textbf{14.20 $\pm$ 1.72} & $40.61 \pm 3.15$ & $47.60 \pm 3.20$ \\
\cline{3-8}
& & \textbf{xPath} & edge 
& $63.60 \pm 2.06$ & $8.80 \pm 2.14$ & $45.93 \pm 1.95$ & $47.60 \pm 1.85$ \\
& & \textbf{GraphSVX} & edge 
& \textbf{93.40 $\pm$ 2.50} & $35.80 \pm 1.94$ & \textbf{54.91 $\pm$ 1.69} & \textbf{55.80 $\pm$ 1.72} \\
& & \textbf{GNNShap} & edge 
& $89.20 \pm 1.94$ & \textbf{44.40 $\pm$ 1.02} & $52.87 \pm 1.28$ & $54.80 \pm 1.60$ \\
& & \textbf{RandomEdgeMask} & edge 
& $66.20 \pm 4.71$ & $8.40 \pm 1.62$ & $48.27 \pm 2.56$ & $50.60 \pm 2.33$\\
\cline{3-8}
& & \textbf{Grad}& node 
& \textbf{82.00 $\pm$ 2.19} & \textbf{35.20 $\pm$ 4.92} & \textbf{49.42 $\pm$ 1.71} & \textbf{49.60 $\pm$ 2.06} \\
& & \textbf{PGM-Explainer}& node 
& $79.40 \pm 3.50$ & $27.60 \pm 2.06$ & $48.70 \pm 2.10$ & $49.00 \pm 2.00$ \\
& & \textbf{RandomNodeMask}& node 
& $63.40 \pm 3.77$ & $11.00 \pm 1.26$ & $44.43 \pm 4.23$ & $49.40 \pm 3.20$ \\
\cline{2-8}
& \multirow{9}{*}{\textbf{HAN-GCN}} 
& \textbf{GNNExplainer} & edge+feature 
& \textbf{65.20 $\pm$ 2.04} & $5.20 \pm 1.33$ & \textbf{42.89 $\pm$ 3.26} & \textbf{48.00 $\pm$ 2.68} \\
& & \textbf{RandomEdgeAndFeatureMask} & edge+feature 
& $60.80 \pm 3.54$ & \textbf{14.00 $\pm$ 3.29} & $36.35 \pm 3.72$ & $45.20 \pm 2.32$ \\
\cline{3-8}
& & \textbf{xPath} & edge 
& $60.00 \pm 5.55$ & $3.80 \pm 1.17$ & $43.29 \pm 2.78$ & $48.80 \pm 1.17$ \\
& & \textbf{GraphSVX} & edge 
& $77.40 \pm 3.38$ & $13.60 \pm 1.96$ & $45.22 \pm 1.07$ & $50.20 \pm 1.17$ \\
& & \textbf{GNNShap} & edge 
& \textbf{78.40 $\pm$ 2.42} & \textbf{23.40 $\pm$ 3.88} & \textbf{47.30 $\pm$ 1.68} & \textbf{51.40 $\pm$ 2.06} \\
& & \textbf{RandomEdgeMask} & edge 
& $67.20 \pm 0.40$ & $6.20 \pm 2.32$ & $42.24 \pm 1.86$ & $48.40 \pm 1.50$\\
\cline{3-8}
& & \textbf{Grad}& node 
& \textbf{89.80 $\pm$ 1.33} & \textbf{33.20 $\pm$ 1.47} & $50.43 \pm 4.82$ & \textbf{52.60 $\pm$ 4.88} \\
& & \textbf{PGM-Explainer}& node 
& $87.20 \pm 2.14$ & $26.00 \pm 3.41$ & \textbf{50.81 $\pm$ 3.16} & $51.20 \pm 2.71$ \\
& & \textbf{RandomNodeMask}& node 
& $68.40 \pm 1.20$ & $9.80 \pm 2.23$ & $43.80 \pm 4.38$ & $49.20 \pm 3.87$\\
\hline
\end{tabular}
\label{tab:explainer_imdb_han}
\end{table*}

\section{Experiments}

We design experiments as a {diagnostic empirical study} of meta-path importance and the reliability of semantic-level meta-path attention.

We organize the study around four research questions:
\begin{enumerate}[leftmargin=*, label=\textbf{RQ\arabic*:}]
    \item How faithful are post-hoc explainers under MetaXplain after being lifted to the meta-path view set?
    \item How much do different meta-path channels contribute to predictive performance under controlled interventions?
    \item Does semantic-level attention reliably reflect meta-path importance?
    \item Can explanations recover task-relevant substructures that are sufficient for prediction?
\end{enumerate}

\subsection{Experiment Setup}

\subsubsection{Evaluation Metrics}
We evaluate explanation quality using the standard faithfulness metrics \textbf{${Fidelity}_{-}$} and \textbf{${Fidelity}_{+}$}~\cite{GraphFramEx}, which quantify {sufficiency} and {necessity} under controlled masking, respectively. For readability, we report \textbf{$1 - {Fidelity}_{-}$} instead of \textbf{${Fidelity}_{-}$}.
To complement local faithfulness with downstream utility, we also report {Macro-F1} and {Micro-F1} on the prediction task under the masked inputs. To standardize comparisons across explainers with different output formats (edge masks, node masks, and edge+feature masks), we evaluate explanations using hard top-$k$ masking at a fixed sparsity level.
Unless otherwise stated, we set sparsity to $0.25$ for each mask type.

\subsubsection{Benchmarks}

We conduct experiments on three benchmark heterogeneous graphs: {ACM}, {DBLP}, and {IMDB}, following the preprocessing and splits in~\cite{gtn} for reproducibility. 
These three datasets all contain two meta-paths. 

\subsubsection{Models}

We evaluate two meta-path-based predictors: {HAN}~\cite{han} and {HAN-GCN}, a HAN variant that replaces GAT layers with GCN layers.
On top of each trained predictor, we apply {MetaXplain} with five representative explainers:
{Grad}~\cite{grad,gradcamgnn}, {GNNExplainer}~\cite{gnnexplainer}, {PGM-Explainer}~\cite{pgm}, {GraphSVX}~\cite{graphsvx}, and {GNNShap}~\cite{gnnshap}.
We compare against the heterogeneous baseline {xPath}~\cite{xpath}, and type-matched random baselines (edge masking, node masking, and edge+feature masking).

\subsection{Effectiveness}

We first address {RQ1} by characterizing how representative post-hoc explainers behave when applied under the
MetaXplain protocol (C1--C3), which standardizes the explanation domain (meta-path views), perturbation semantics, and
evaluation budget. We provide a controlled comparison
across explanation families and identify robust patterns and failure modes that matter for subsequent attention--importance diagnostics.

Table~\ref{tab:explainer_imdb_han} report faithfulness and downstream performance
on {ACM}, {DBLP}, and {IMDB} for both {HAN} and {HAN-GCN} over 5 random seeds.
To isolate explanation quality from mask budget, we compare methods within each \textbf{mask type}
(edge+feature, edge, node) against a type-matched random baseline at the same sparsity.

\paragraph{Overall effectiveness (RQ1).}
Across datasets and backbones, most MetaXplain-applied explainers improve over their corresponding random controls,
indicating that view-factorized explanations recover non-trivial, prediction-relevant evidence.
However, the magnitude and stability of these improvements are strongly \emph{backbone-dependent} and vary across explainer families.

\paragraph{Edge+feature masking: GNNExplainer is backbone-sensitive.}
{GNNExplainer} performs strongly on {HAN} (e.g., consistent improvements over
\textit{RandomEdgeAndFeatureMask} across datasets), but can degrade sharply on {HAN-GCN}.
We attribute this to a {soft-mask vs.\ hard-mask gap}: GNNExplainer optimizes continuous reweighting, but evaluation
uses hard top-$k$ sparsification. This discretization-induced shift is amplified under GCN-style aggregation and
normalization, making HAN-GCN explanations brittle compared to HAN.

\paragraph{Edge masking: Shapley-style explainers are comparatively robust.}
xPath can be competitive in some settings, but its behavior is not consistently superior to type-matched random controls across datasets/backbones.
In contrast, Shapley-style explainers (GraphSVX and GNNShap) tend to be more robust: they frequently exceed both xPath and \textit{RandomEdgeMask},
and show smaller sensitivity to the choice of backbone.
Moreover, GNNShap often improves over GraphSVX, consistent with edge-level Shapley attribution being better aligned with edge-masking evaluation
than surrogate node-centric approximations~\cite{gnnshap}.

\paragraph{Node masking: Grad is consistently strong.}
Within node-masking methods, Grad is typically the strongest and substantially exceeds \textit{RandomNodeMask}.
PGM-Explainer also improves over the random control, but is generally less competitive than Grad.
This gap suggests headroom for methods that more directly model meta-path-specific evidence under heterogeneous semantics,
motivating our later focus on meta-path contribution estimation and attention alignment (RQ2--RQ3).

\paragraph{Backbone sensitivity: HAN vs.\ HAN-GCN}
Finally, we observe higher variance and stronger method sensitivity on {HAN-GCN} than on {HAN}
(e.g., larger standard deviations for several edge explainers on ACM/HAN-GCN).
This indicates that the underlying within-view aggregation and cross-view fusion mechanisms play a central role in explanation stability,
and motivates our subsequent diagnostic analysis of semantic attention and meta-path importance (RQ3).

\begin{table}[ht]
\centering
\small
\caption{MP-AEA scores across 100 random seeds.}

\begin{tabular}{l|c|c|c}
\hline

\textbf{Metric}             & \textbf{ACM} & \textbf{DBLP} & \textbf{IMDB} \\
\hline
\multicolumn{4}{c}{\textbf{HAN}} \\
\hline
\textbf{Kendall’s tau (P-value)}    & 0.439 (1e-10) & 0.053 (0.432) & 0.700 (6e-25) \\
\textbf{Spearman (P-value)}          & 0.618 (8e-12) & 0.069 (0.496) & 0.872 (4e-32) \\
\hline
\multicolumn{4}{c}{\textbf{HAN-GCN}} \\
\hline
\textbf{Kendall’s tau (P-value)}    & 0.414 (1e-9) & 0.601 (8e-19) & 0.860 (8e-37) \\
\textbf{Spearman (P-value)}          & 0.562 (1e-9) & 0.793 (8e-23) & 0.961 (1e-56) \\
\hline
\end{tabular}
\label{tab:align_han}

\vspace{-1em}
\end{table}

\begin{figure*}[ht]
    \centering
    \includegraphics[scale=0.35]{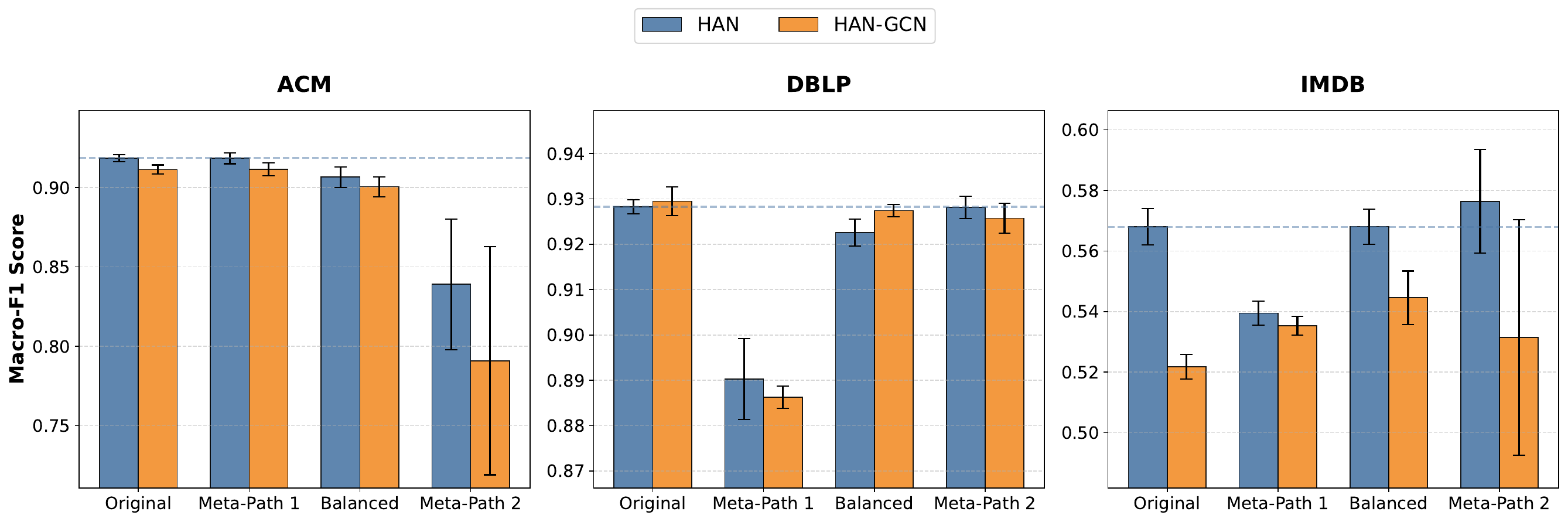}

    \caption{\textbf{Macro-F1 under controlled semantic attention interventions.}
        Error bars show std.\ over 5 runs.}
    \label{macroAtten_fig}
\end{figure*}
\begin{figure}[ht]
    \centering
    \includegraphics[scale=0.25]{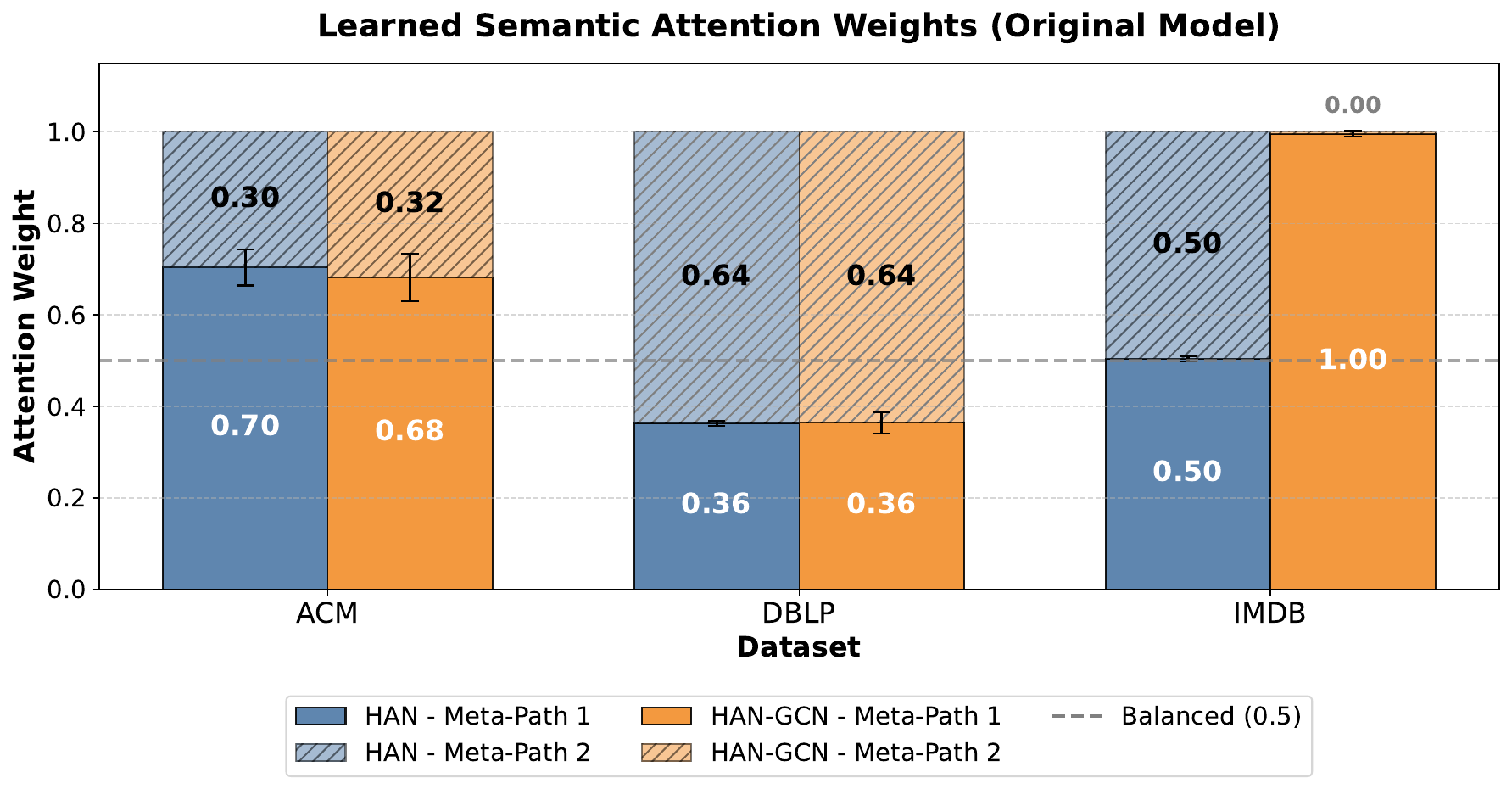}

    \caption{\textbf{Learned semantic attention weights.}
        Error bars indicate std.}
    \label{oriAtten_fig}
    \vspace{-1em}
\end{figure}

\subsection{Attention can decouple from meta-path importance (RQ2--RQ3)}
\label{sec:mp_importance}

This section addresses the central diagnostic question of our study: \textbf{when does semantic-level meta-path attention reflect meta-path importance, and when can it decouple?}
We probe meta-path importance using explanation-derived per-view contribution scores (Section~\ref{sec:mp_aea}) and quantify attention reliability using \textbf{MP-AEA}, which measures rank correlation between learned semantic attention and explanation-derived meta-path contributions across random runs.
In addition, we perform \emph{controlled attention interventions} (forcing attention to a single meta-path or to a uniform mixture) to test the empirical utility of each meta-path channel.

\emph{Why HAN-GCN is a cleaner probe of meta-path importance.}
HAN couples {edge-level} attention within each meta-path view (GAT) with {semantic-level} attention across meta-path channels, so strong performance may come from neighbor selection within a view even when a channel is down-weighted. Prior works~\cite{hgnnbench,autognr} show that well-tuned attention-based GNNs (e.g., GAT) can be highly competitive on heterogeneous benchmarks, suggesting edge attention can partially substitute for explicit heterogeneous semantics. We therefore also evaluate HAN-GCN, which removes within-view edge attention and makes semantic attention the primary mechanism for weighting meta-path channels. This separation helps interpret whether attention behaves as a proxy for channel importance rather than as one component among multiple attention mechanisms.

\subsubsection{Does Attention Align with Importance? (RQ3)}
We quantify the agreement between semantic attention and explanation-derived meta-path contributions using
MP-AEA (Section~\ref{sec:mp_aea}). Tables~\ref{tab:align_han} report Kendall's $\tau$ and
Spearman's $\rho$ correlations between (i) semantic attention weights and (ii) Grad-based meta-path contribution
scores across 100 random seeds.

\emph{Key observation: both alignment and decoupling regimes exist.}
On IMDB, both HAN and HAN-GCN exhibit consistently high correlations, indicating that semantic attention is a stable proxy for explanation-derived meta-path importance in this setting.
In contrast, on DBLP the correlation for HAN is statistically insignificant, revealing a clear \emph{decoupling} regime where attention does not reliably track meta-path contribution estimates.
Notably, alignment is generally stronger under HAN-GCN than HAN across datasets, consistent with HAN-GCN isolating semantic attention from within-view edge attention.

\subsubsection{Empirical Utility of Meta-Paths (RQ2)}
We next test whether meta-paths carry materially different predictive signal via controlled semantic attention interventions
(Figures~\ref{macroAtten_fig} and~\ref{microAtten_fig} in Appendix). Specifically, we compare the learned attention
(\textit{Original}) with three controlled configurations: forcing attention to a single meta-path
(\textit{Meta-Path 1} or \textit{Meta-Path 2}) and a uniform mixture (\textit{Balanced}).

\emph{Meta-path utility is dataset-dependent and can expose attention failure modes.}
On ACM and DBLP, performance is strongly sensitive to path selection: forcing attention to the empirically stronger meta-path nearly matches the learned model, while forcing the weaker meta-path substantially degrades performance.
This indicates that the two semantic channels contain materially different task-relevant signal.
On IMDB, we observe a distinct behavior: under HAN-GCN, semantic attention collapses toward a single meta-path, yet the \textit{Balanced} configuration performs better.
This suggests that attention can over-commit to one channel and miss complementary evidence, even when both meta-paths are useful under controlled mixing.

\emph{Connection to subgraph recoverability.}
Importantly, a decoupling or over-commitment regime does not imply that meta-path signal is absent.
As shown in Section~\ref{sec:subgraph_fidelity}, retraining on explanation-induced subgraphs can substantially improve IMDB performance for HAN-GCN, consistent with an explanation-as-denoising effect in noisy regimes.
Together, these results support our main message: \emph{semantic attention is not universally reliable as a proxy for meta-path importance}, and its reliability depends on the dataset and backbone.

\begin{figure}[ht]
    \centering
    \includegraphics[width=0.47\textwidth]{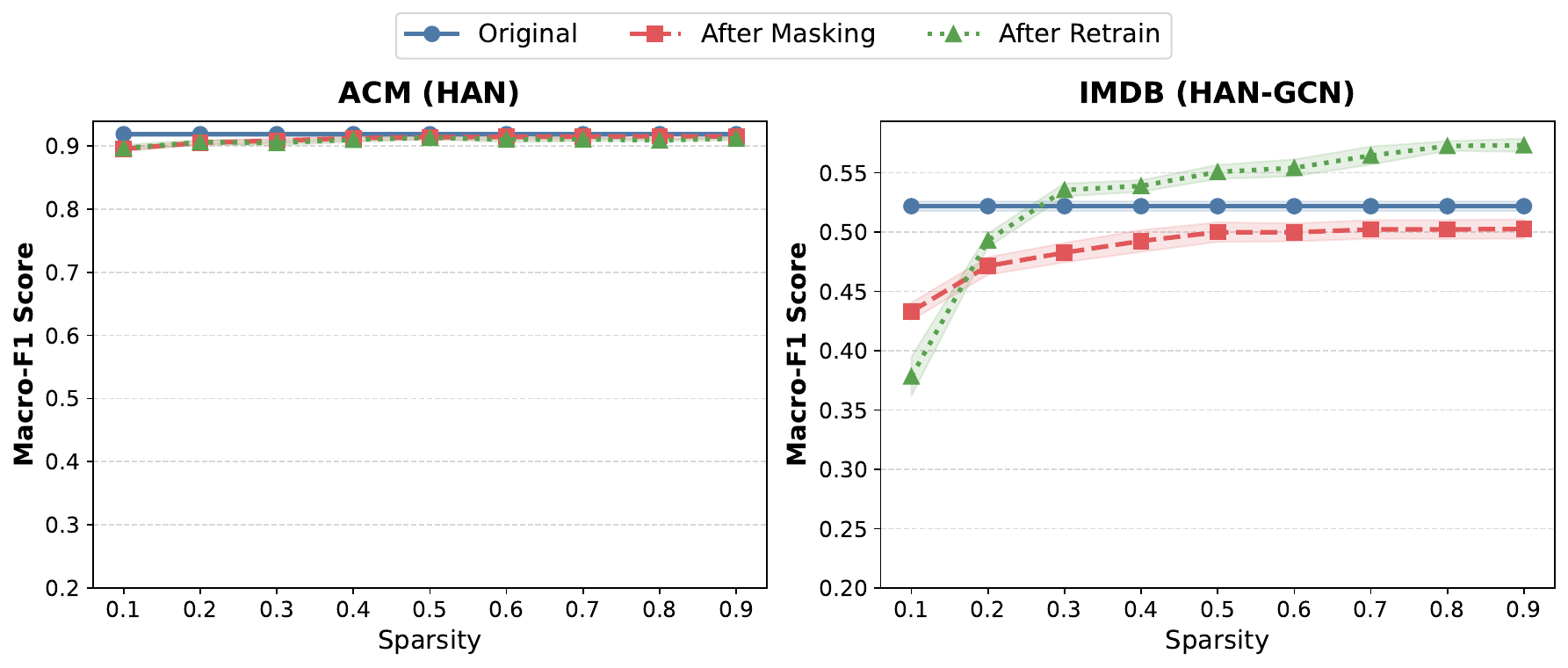}

    \caption{\textbf{Recoverability from explanation-induced subgraphs.}
        Shaded regions show std.\ over 5 runs.}
    \label{retrain_small}
\end{figure}

\subsection{Recoverability from explanation-induced subgraphs (RQ4)}
\label{sec:subgraph_fidelity}

To address \textbf{RQ4}, we use explanation-induced subgraph extraction as a \emph{diagnostic intervention}:
if a post-hoc explanation identifies task-relevant evidence, then the subgraph induced by top-ranked nodes should retain
substantial predictive signal.
Importantly, this test separates two notions:
(i) \emph{inference-time sufficiency} under hard pruning (which can introduce distribution shift), and
(ii) \emph{recoverability} of the signal after retraining on the extracted structure.

\emph{Protocol.}
For each target node $v$, we construct an explanation-induced subgraph by retaining the top-ranked nodes under {Grad}
at a prescribed sparsity level (fraction removed).
We then compare three settings:
\emph{Original} (performance on the full graph),
\emph{After Masking} (inference on the extracted subgraph without retraining), and
\emph{After Retrain} (training a new model on the extracted subgraph with the same hyperparameters).
Figure~\ref{retrain_small} reports Macro-F1 under varying sparsity.

\emph{Case (i): ACM with HAN --- concentrated evidence and mild distribution shift.}
On ACM/HAN, \emph{After Retrain} remains close to \emph{Original} even at high sparsity, indicating that the
task-relevant evidence is concentrated in a small neighborhood that the explainer consistently recovers.
Moreover, the relatively small gap between \emph{After Masking} and \emph{Original} suggests that the extracted subgraphs
remain largely in-distribution for the trained attention-based backbone, so pruning does not severely disrupt inference.

\emph{Case (ii): IMDB with HAN-GCN --- recoverability and an explanation-as-denoising regime.}
On IMDB/HAN-GCN, hard pruning causes a clear drop in \emph{After Masking}, reflecting sensitivity of GCN-style aggregation
to structural removal and potential distribution shift.
However, \emph{After Retrain} substantially recovers this drop and can even exceed \emph{Original} at moderate-to-high sparsity.
This behavior is consistent with an \emph{explanation-as-denoising} regime: pruning weakly relevant neighborhoods can act as
a beneficial inductive bias for retraining by reducing noise in the computation graph.

\emph{Takeaway.}
These results support a key theme of our study: even when semantic attention can decouple from meta-path importance (RQ3),
post-hoc explanations can still identify substructures that retain recoverable predictive signal.
The gap between \emph{After Masking} and \emph{After Retrain} further highlights that evaluating explanations solely by
inference-time masking can conflate explanation quality with distribution shift, especially for GCN-style backbones.

\section{Conclusion}
We conducted an empirical study of a common interpretability assumption in meta-path-based heterogeneous GNNs: whether semantic-level meta-path attention can be treated as a proxy for meta-path importance. To make this question testable without changing the predictor, we introduced \textbf{MetaXplain}, a meta-path-consistent post-hoc explanation protocol that lifts existing explainers into the meta-path view domain via view-factorized explanations, schema-valid channel-wise perturbations, and fusion-aware attribution, and we proposed \textbf{MP-AEA} to quantify attention--\\importance agreement across random runs. Experiments on ACM, DBLP, and IMDB with HAN and HAN-GCN show that lifted explainers generally outperform type-matched random baselines but are backbone-sensitive, while MP-AEA reveals both alignment and decoupling regimes where attention can or cannot reliably track explanation-derived meta-path contributions depending on the dataset/backbone. Finally, explanation-induced subgraphs often retain recoverable predictive signal after retraining and can even improve performance in noisy regimes, suggesting an explanation-as-denoising effect and highlighting that masking-only evaluation can conflate explanation quality with distribution shift.

\bibliographystyle{ACM-Reference-Format}
\bibliography{ref}
\appendix

\section{Additional Results}

\begin{figure*}[ht]
    \centering
    \includegraphics[scale=0.35]{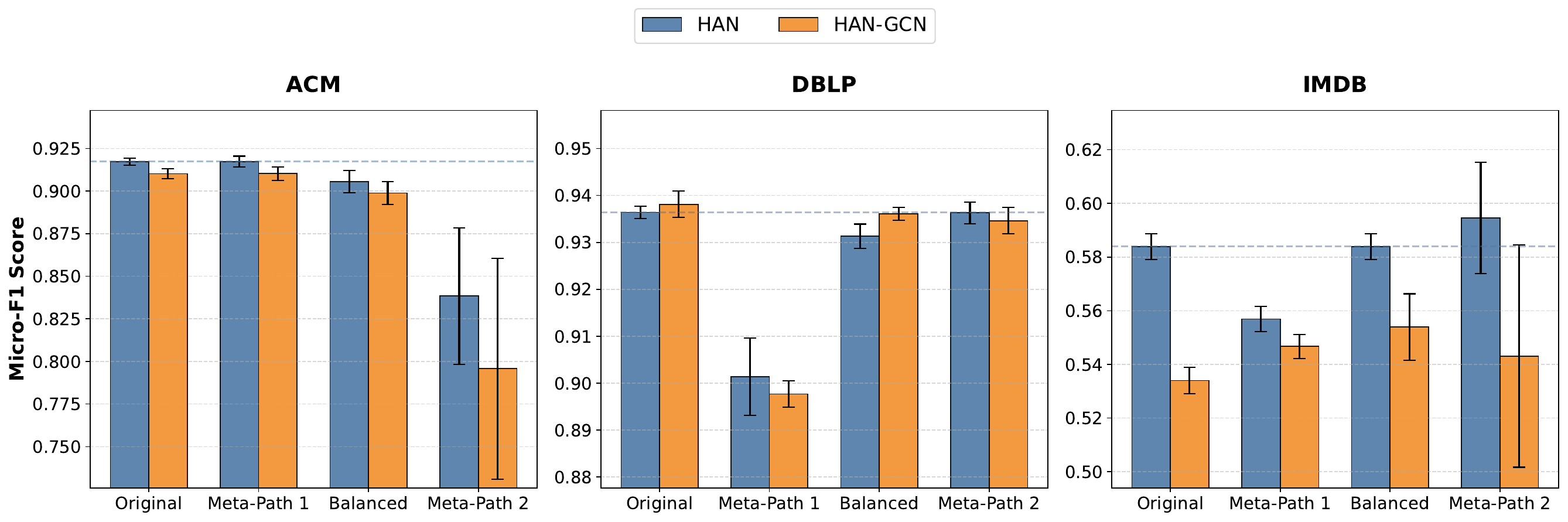}

    \caption{\textbf{Micro-F1 under controlled semantic attention interventions.}
        Error bars show std.\ over 5 runs.}
    \label{microAtten_fig}
\end{figure*}

\begin{figure*}[ht]
    \centering
    \includegraphics[scale=0.4]{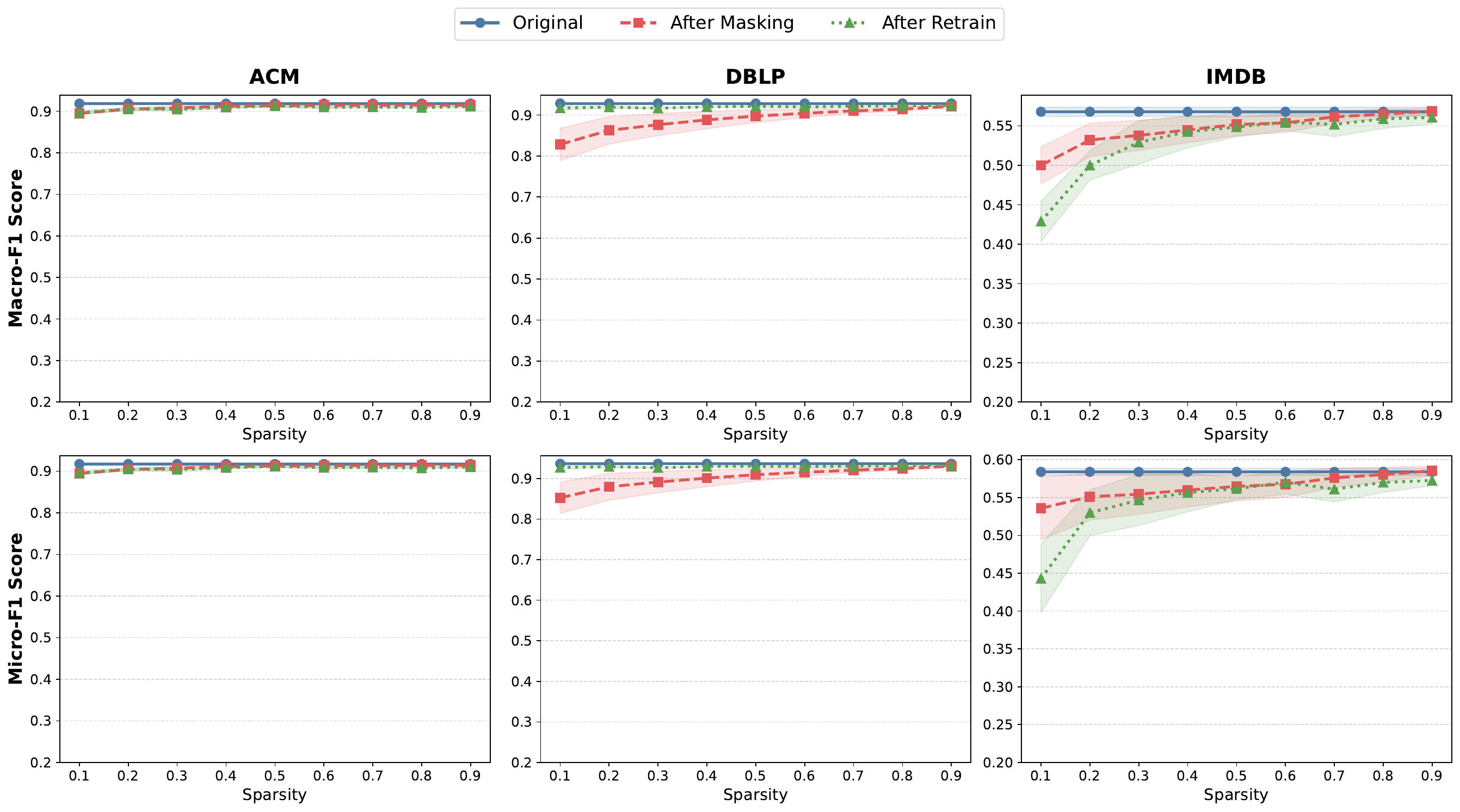}

    \caption{\textbf{Recoverability from explanation-induced subgraphs.}
        We evaluate the quality of explanation-identified subgraphs at varying sparsity levels (fraction of nodes removed).
        Three conditions are compared:
        \textit{Original} (full graph performance),
        \textit{After Masking} (inference on the extracted subgraph without retraining), and
        \textit{After Retrain} (model retrained on the extracted subgraph).
        Shaded regions indicate std. over 5 runs.}
    \label{retrain_HAN}
\end{figure*}

\begin{figure*}[ht]
    \centering
    \includegraphics[scale=0.4]{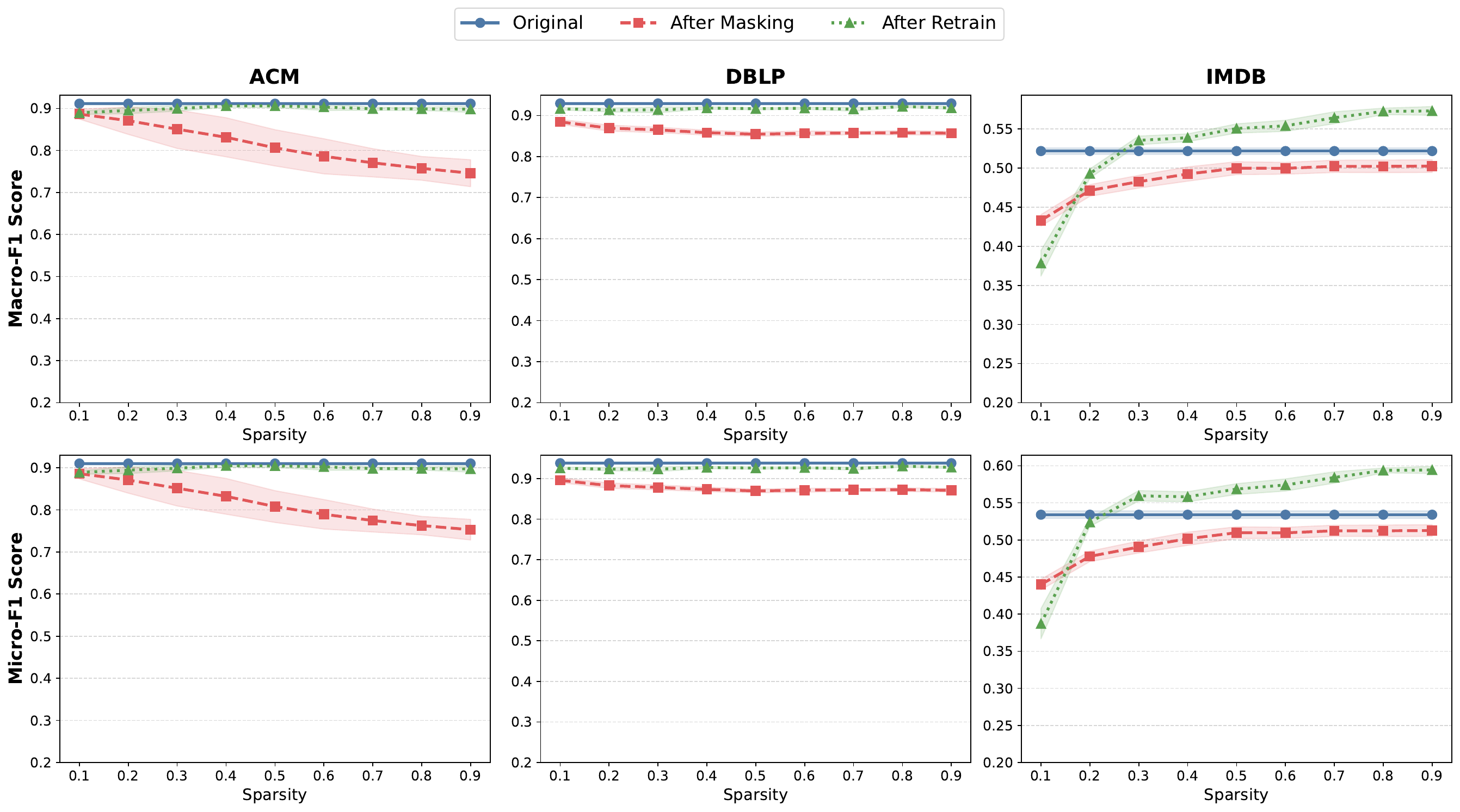
    }

    \caption{\textbf{Recoverability from explanation-induced subgraphs.}
        Shaded regions indicate std. over 5 runs.}
    \label{retrain_HAN_GCN}
\end{figure*}

\subsection{Additional Analysis for Subgraph Fidelity}

\paragraph{HAN: extracted subgraphs are largely sufficient on citation-style benchmarks.}
Across ACM and DBLP (Figure~\ref{retrain_HAN}), \emph{After Retrain} stays close to \emph{Original} even at high sparsity,
suggesting that predictive evidence is concentrated and that \textsc{Grad} consistently recovers the most informative neighborhood.
In these settings, \emph{After Masking} also remains relatively stable, indicating that the pruned computation graph does not
introduce severe distribution shift for an attention-based aggregator.

\paragraph{HAN-GCN: sensitivity to structural perturbations, but strong recoverability after retraining.}
For HAN-GCN on ACM (Figure~\ref{retrain_HAN_GCN}), \emph{After Masking} degrades more consistently with increasing sparsity compared
to HAN, reflecting the greater sensitivity of a GCN-style aggregator to hard structural changes.
However, \emph{After Retrain} remains near \emph{Original}, indicating that the pruned graphs still contain sufficient predictive
signal and that the model can relearn effectively on the explanation-induced structure.

\section{Datasets}

We evaluate MetaXplain on three widely used heterogeneous graph benchmarks for node classification: ACM, DBLP, and IMDB. These datasets differ in domain, node semantics, and relational structure, allowing us to assess explanation behavior across diverse heterogeneous settings. Dataset statistics are summarized in Table~\ref{tab:dataset_stats}.

\paragraph{ACM}
ACM is a citation network with three node types:
\emph{papers (P)}, \emph{authors (A)}, and \emph{subjects (S)}.
Edges include paper--author (P--A), paper--subject (P--S), and paper--paper citations.
The node classification task is to predict the research category of papers.
Each paper is associated with a bag-of-words feature vector.
Meta-paths such as P--A--P and P--S--P are used to model collaboration and topic-based semantics.

\paragraph{DBLP}
DBLP is a bibliographic network consisting of three node types:
\emph{authors (A)}, \emph{papers (P)}, and \emph{conferences (C)}.
Edges encode relations such as author--paper (A--P), and paper--conference (P--C).
The task is to classify \emph{author} nodes into research areas.
Node features are represented as bag-of-words vectors derived from paper titles.
We construct 2 meta-path views (e.g., A--P--A, A--P--C--P--A) to capture
distinct semantic relations between authors.

\paragraph{IMDB}
IMDB is a movie network containing \emph{movies (M)}, \emph{actors (A)}, and \emph{directors (D)}.
Edges represent acting and directing relationships.
The task is multi-class movie genre classification.
Node features are constructed from plot keywords.
We define meta-paths such as M--A--M and M--D--M to encode shared cast and director information.

\begin{table}[ht]
\centering
\caption{Statistics of heterogeneous graph datasets used for node classification.}
\label{tab:dataset_stats}
\setlength{\tabcolsep}{3pt}
\begin{tabular}{lccccccc}
\hline
\textbf{Dataset} &
\textbf{\#Nodes} &
\textbf{\#Edges} &
\textbf{\#$\mathcal{R}$} &
\textbf{\#Feat.} &
\textbf{Train} &
\textbf{Val} &
\textbf{Test} \\
\hline
ACM  & 8{,}994  & 25{,}922 & 4 & 1{,}902 & 600 & 300 & 2{,}125 \\
DBLP & 18{,}405 & 67{,}946 & 4 & 334  & 800 & 400 & 2{,}857 \\
IMDB & 12{,}772 & 37{,}288 & 4 & 1{,}256 & 300 & 300 & 2{,}339 \\
\hline
\end{tabular}

\end{table}

\section{Model Architectures and Training Hyperparameters}
All heterogeneous GNN models are trained prior to explanation using fixed architecture and optimization settings, based on the original authors’ reported hyperparameter settings. This section summarizes the shared training hyperparameters for HAN and HAN-GCN models used across all datasets.

\begin{table}[ht]
\centering
\caption{Training hyperparameters for heterogeneous GNN models.}
\label{tab:model_hparams}
\setlength{\tabcolsep}{6pt}
\begin{tabularx}{\columnwidth}{Xcc}
\hline
\textbf{Hyperparameter} & \textbf{HAN} & \textbf{HAN\_GCN} \\
\hline
Node-level aggregation & GAT & GCN \\
Hidden units & 8 & 64 \\
Attention heads & 8 & -- \\
Dropout rate & 0.6 & 0.6 \\
Learning rate & 0.01 & 0.01 \\
Weight decay & 0.001 & 0.001 \\
Optimizer & Adam & Adam \\
Training epochs & 200 & 200 \\
Early stopping patience & 100 & 100 \\
\hline
\end{tabularx}
\end{table}

\section{More Information about Extended Algorithms under MetaXplain}

\subsection{Perturbation-Based Meta-Path Explanations}

\textbf{PGM-Explainer.}
Under C1--C2, we apply PGM-Explainer to produce meta-path-specific node importances via randomized perturbations and conditional dependence testing.
For each meta-path view $m$, we collect candidate nodes in the extracted neighborhood and generate binary perturbation
samples by randomly perturbing node features (independently per view) and evaluating the predictor.
We then compute a $\chi^2$ contingency test between the perturbation state of each candidate node and the perturbation
sensitivity of the target node prediction, yielding a p-value per node per view.
Finally, we convert p-values into importance weights (smaller p-values $\Rightarrow$ larger weights), producing a node
mask vector for each meta-path view.

\subsection{Shapley-Style Meta-Path Explanations}
Shapley-style explainers treat structural elements as players in a cooperative game.
Under C3(B), we compute Shapley attributions conditioned on a single meta-path view while holding the other views fixed.

\textbf{GNNShap.}
For a chosen meta-path view $m$ with $E_m$ edges, we define a coalition by a binary vector
$\mathbf{z}\in\{0,1\}^{E_m}$ and form $G^{(m)}(\mathbf{z})$ by removing edges with $z_e=0$.
The value function evaluates the predictor with only view $m$ masked:
\begin{equation}
v_m(\mathbf{z}) =
f\!\left(G^{(1)}_v,\dots,G^{(m)}_v(\mathbf{z}),\dots,G^{(M)}_v; v\right).
\end{equation}
We then fit a weighted linear surrogate (KernelSHAP-style) over sampled coalitions to estimate edge Shapley values
$\{\mathcal{S}_e\}_{e=1}^{E_m}$, producing edge attributions per view.

\textbf{GraphSVX.}
GraphSVX defines players as neighboring nodes of $v$ within a fixed meta-path view, samples node coalitions, and fits a
KernelSHAP-style linear surrogate to obtain node attributions.
In our pipeline, node-level attributions can be optionally lifted to edge scores by assigning each node’s value to its incident edges for comparison against edge-level explainers.


\begin{algorithm}[ht]
\caption{Grad under MetaXplain}
\label{alg:mp_grad}
\begin{algorithmic}[1]
\STATE \textbf{Input:} Heterogeneous graph $G$, trained HeteroGNN $f$, meta-path set $\mathcal{M}$, target node $v$, hops $k$
\STATE \textbf{Output:} Meta-path–conditioned node importance $\{I_{\mathrm{Grad}}^{(m)}\}_{m \in \mathcal{M}}$

\STATE Extract $k$-hop subgraph $\mathcal{G}(v)$

\STATE Initialize feature tensors $\{X^{(m)}\}$ with gradient tracking
\STATE Compute prediction $f(v)$ on $\mathcal{G}(v)$
\STATE $c^\star \leftarrow \arg\max f(v)$
\STATE Compute loss $\mathcal{L}(v, c^\star)$

\FOR{$m \in \mathcal{M}$}
    \STATE $\nabla_{X^{(m)}} \mathcal{L} \leftarrow \partial \mathcal{L} / \partial X^{(m)}$
    \STATE $I_{\mathrm{Grad}}^{(m)}(i) \leftarrow \left\lVert \partial \mathcal{L} / \partial x_i^{(m)} \right\rVert_2$
\ENDFOR

\STATE \textbf{return} $\{I_{\mathrm{Grad}}^{(m)}\}_{m \in \mathcal{M}}$
\end{algorithmic}
\end{algorithm}

\begin{algorithm}[ht]
\caption{GNNExplainer under MetaXplain}
\label{alg:mp_gnnexplainer}
\begin{algorithmic}[1]
\STATE \textbf{Input:} Heterogeneous graph $G$, trained HeteroGNN $f$, meta-path set $\mathcal{M}$,
target node $v$, hops $k$, optimization steps $T$
\STATE \textbf{Output:} Meta-path–conditioned edge masks $\{M_A^{(m)}\}_{m \in \mathcal{M}}$ and shared feature mask $M_X$

\STATE Extract $k$-hop subgraph $\mathcal{G}(v)$
\STATE Initialize edge mask $M_A^{(m)}$ for each meta-path and shared feature mask $M_X$
\FOR{$t = 1$ to $T$}
    \FOR{$m \in \mathcal{M}$}
    
        \STATE $\widetilde{A}^{(m)} \leftarrow A^{(m)} \odot \sigma(M_A^{(m)})$
        \STATE $\widetilde{X} \leftarrow X \odot \sigma(M_X)$
        \STATE $\hat{y}^{(m)} \leftarrow f(\widetilde{A}^{(m)}, \widetilde{X})$
    \ENDFOR
    \STATE Compute $\mathcal{L}$ by Equation~\ref{L_GNNExplainer}
    \STATE Update $(\{M_A^{(m)}\}, M_X)$ via gradient descent
\ENDFOR

\STATE \textbf{return} $\{M_A^{(m)}\}_{m \in \mathcal{M}},\; M_X$
\end{algorithmic}
\end{algorithm}


\begin{table}[ht]
\centering
\caption{Hyperparameters for GNNExplainer.}
\label{tab:gnnexplainer_hparams}
\begin{tabularx}{\columnwidth}{lX}
\hline
\textbf{Hyperparameter} & \textbf{Value} \\
\hline
Optimization epochs & 100 \\
Optimizer & Adam with lr = 0.1 \\
Prediction loss coefficient ($\lambda_{\text{pred}}$) & 1.0 \\
Edge sparsity coefficient
($\lambda_e$) & 0.005 \\
Feature sparsity coefficient ($\lambda_x$) & 1.0 \\
Entropy coefficient ($\lambda_{\text{ent}}$) &  edge: 1.0; feature: 0.1 \\
Laplacian smoothness coefficient ($\lambda_{\text{lap}}$) & 1.0 \\
\hline
\end{tabularx}
\end{table}


\begin{algorithm}[ht]
\caption{PGM-Explainer under MetaXplain}
\label{alg:mp_pgmexplainer}
\begin{algorithmic}[1]
\STATE \textbf{Input:} Heterogeneous graph $G$, trained HeteroGNN $f$, meta-path set $\mathcal{M}$,
target node $v$, hops $k$, perturbation samples $n$
\STATE \textbf{Output:} Meta-path–conditioned probabilistic explanations
$\{\mathcal{B}^{(m)}\}_{m \in \mathcal{M}}$

\STATE Extract $k$-hop subgraph $\mathcal{G}(v)$

\FOR{$m \in \mathcal{M}$}
    \STATE Identify node set $\mathcal{N}_v^{(m)}$
    
    \STATE Initialize perturbation dataset $\mathcal{D}^{(m)}$
    \FOR{$i = 1$ to $n$}
        \STATE Perturb node features in $\mathcal{N}_v^{(m)}$
        \STATE Evaluate $f(\mathcal{G}(v))$ with only view $m$ perturbed
        \STATE Append perturbation and prediction change to $\mathcal{D}^{(m)}$
    \ENDFOR
\ENDFOR
\STATE Perform conditional independence tests on $\{\mathcal{D}^{(m)}\}$
\STATE Select statistically dependent nodes from each meta-path
\STATE Learn Bayesian network $\{\mathcal{B}^{(m)}\}$

\STATE \textbf{return} $\{\mathcal{B}^{(m)}\}_{m \in \mathcal{M}}$
\end{algorithmic}
\end{algorithm}


\begin{table}[ht]
\centering
\caption{Hyperparameters for PGM-Explainer.}
\label{tab:pgmexplainer_hparams}
\begin{tabular}{lc}
\hline
\textbf{Hyperparameter} & \textbf{Value} \\
\hline
Perturbation samples & 10 \\
$\chi^2$ threshold & 0.05 \\
Prediction drop threshold & 0.1 \\
Feature perturbation mode & Binary \\
\hline
\end{tabular}
\end{table}


\begin{algorithm}[ht]
\caption{GNNShap under MetaXplain}
\label{alg:mp_gnnshap}
\begin{algorithmic}[1]
\STATE \textbf{Input:} Heterogeneous graph $G$, trained HeteroGNN $f$, meta-path set $\mathcal{M}$,
target node $v$, hops $k$, coalition samples $n$
\STATE \textbf{Output:} Meta-path–conditioned Shapley values
$\{\mathcal{S}\}_{m \in \mathcal{M}}$

\STATE Extract $k$-hop subgraphs  $\mathcal{G}(v)$
\STATE Compute full prediction $(f_{\mathrm{full}}, c^\star) \leftarrow f(G_v, v)$

\FOR{$m \in \mathcal{M}$}
    \STATE Let $E^{(m)} \leftarrow |E(G_v^{(m)})|$
    \IF{$E^{(m)} \leq 1$}
        \STATE $\boldsymbol{\phi}^{(m)} \leftarrow \mathbf{0}$
        \STATE \textbf{continue}
    \ENDIF

    \STATE Build edge-player set $\mathcal{P}^{(m)} \leftarrow E(G_v^{(m)})$
    \STATE Instantiate sampler $\mathcal{S}^{(m)}$ with $|\mathcal{P}^{(m)}|$ players and $n$ samples
    \STATE Sample coalition masks $Z^{(m)}$ and kernel weights $\boldsymbol{\pi}^{(m)}$

    \STATE Compute null prediction $f_{\mathrm{null}}^{(m)}$ (meta-path removed)
    \STATE Compute coalition predictions 
    \[\hat{y}^{(m)} \leftarrow f(\{G_v^{(i)}|i\neq m\}, G_v^{(m)}(z); v)\]

    \STATE Select solver:
    \[
    \text{WLS if } E^{(m)} \leq 1000,\quad \text{else WLR}
    \]

    \STATE Solve weighted regression to obtain $\mathcal{S}^{(m)}$
\ENDFOR

\STATE \textbf{return} $\{\mathcal{S}^{(m)}\}_{m \in \mathcal{M}}$
\end{algorithmic}
\end{algorithm}


\begin{table}[ht]
\centering
\caption{Hyperparameters for GNNShap.}
\label{tab:gnnshap_hparams}
\begin{tabular}{lc}
\hline
\textbf{Hyperparameter} & \textbf{Value} \\
\hline
Coalition samples & 15{,}000 \\
Shapley solver & WLS ($E \leq 1000$), WLR (otherwise) \\
Ridge regularization & $10^{-3}$ \\
\hline
\end{tabular}
\end{table}


\begin{algorithm}[ht]
\caption{GraphSVX under MetaXplain}
\label{alg:mp_graphsvx}
\begin{algorithmic}[1]
\STATE \textbf{Input:} Heterogeneous graph $G$, trained HeteroGNN $f$, meta-path set $\mathcal{M}$,
target node $v$, hops $k$, coalition samples $n$, size parameter $S$
\STATE \textbf{Output:} Meta-path–conditioned Shapley values
$\{\mathcal{S}^{(m)}\}_{m \in \mathcal{M}}$

\STATE Extract $k$-hop subgraphs  $\mathcal{G}(v)$
\STATE Map target node $v$ to local index $v^\prime$

\STATE Compute full prediction $(p^\star, c^\star) \leftarrow f(G_v, v^\prime)$
\STATE Select relevant feature indices $\mathcal{F}$

\FOR{$m \in \mathcal{M}$}
    \STATE Identify neighbor set $\mathcal{N}_v^{(m)}$
    \STATE Determine number of players $M^{(m)} \leftarrow |\mathcal{N}_v^{(m)}| + |\mathcal{F}|$
    
    \IF{$M^{(m)} \leq 1$}
        \STATE $\boldsymbol{\phi}^{(m)} \leftarrow \mathbf{0}$
        \STATE \textbf{continue}
    \ENDIF

    \STATE Sample coalition masks $Z^{(m)} \in \{0,1\}^{n \times M^{(m)}}$
    \STATE Compute Shapley kernel weights $\boldsymbol{\pi}^{(m)}$

    \FOR{each coalition $z_i \in Z^{(m)}$}
        \STATE Mask neighbors and features according to $z_i$
        \STATE Evaluate masked prediction $f_i^{(m)} \leftarrow f(G_v^{(m)}, z_i)$
    \ENDFOR

    \STATE Fit weighted linear model $g^{(m)}$ on $(Z^{(m)}, f^{(m)})$
    \STATE Extract Shapley values $\mathcal{S}^{(m)}$ from $g^{(m)}$
\ENDFOR

\STATE \textbf{return} $\{\mathcal{S}^{(m)}\}_{m \in \mathcal{M}}$
\end{algorithmic}
\end{algorithm}


\begin{table}[ht]
\centering
\caption{Hyperparameters for GraphSVX.}
\label{tab:graphsvx_hparams}
\begin{tabular}{lc}
\hline
\textbf{Hyperparameter} & \textbf{Value} \\
\hline
Coalition samples & 1000 \\
Maximum coalition size & 3 \\
Coalition sampling strategy & SmarterSeparate \\
Feature marginalization & Expectation \\
Surrogate model & Weighted Linear Regression \\
Node--feature balance & nodes only \\
Multi-class explanation & False \\
\hline
\end{tabular}
\end{table}

\section{xPath Implementation Details}

We set the sampling budget to $m{=}10$ and the candidate budget to $b{=}5$ in all experiments.

\end{document}